\documentclass[a4paper,fleqn]{cas-dc}

\usepackage{setspace}
\usepackage{amsmath,amssymb,amsfonts}

\usepackage{bbm} 
\usepackage{bm}
\usepackage{algorithmic}
\usepackage{algorithm}
\usepackage{tikz}
\usepackage{comment}
\usepackage{color}
\usepackage{overpic}
\usepackage{graphicx}
\usepackage{pifont}
\usepackage{pict2e} 
\usepackage{booktabs}
\usepackage{multirow}
\usepackage{multicol}
\usepackage{xcolor,colortbl}
\usepackage{nicematrix}

\definecolor{Gray}{gray}{0.95}
\definecolor{darkgreen}{rgb}{0,0.5,0}
\definecolor{LightCyan}{rgb}{0.88,1,1}
\definecolor{LightRed}{rgb}{1,0.9,0.9}

\usepackage[capitalize,noabbrev]{cleveref}
\crefname{section}{Sec.}{Secs.}
\crefname{figure}{Fig.}{Figs.}
\crefname{table}{Tab.}{Tabs.}
\crefname{equation}{Eq.}{Eqs.}
\crefname{algorithm}{Alg.}{Algs.}

\definecolor{color1}{RGB}{237,125,49}
\definecolor{color3}{RGB}{0,112,192}
\definecolor{color2}{RGB}{112,173,71}
\usepackage{etoolbox}
\usepackage{listofitems}
\usepackage{wheelchart}
\usetikzlibrary{decorations.text}
\usetikzlibrary{decorations.markings}
\usepackage{makecell}
\usepackage{wrapfig}

\usepackage{forest}
\usetikzlibrary{trees,positioning,shapes,shadows,arrows.meta}

\makeatletter
\renewcommand\subsubsection{\@startsection{subsubsection}{4}{1\parindent}{0ex plus 0.1ex minus 0.1ex}{0ex}{\normalfont\normalsize\bfseries\itshape}}\renewcommand\paragraph{\@startsection{paragraph}{4}{2\parindent}{0ex plus 0.1ex minus 0.1ex}{0ex}{\normalfont\normalsize\bfseries\itshape}}\makeatother

\usetikzlibrary{trees}
\definecolor{hidden-draw}{rgb}{0.5, 0.5, 0.5}
\definecolor{harvestgold}{rgb}{0.85, 0.57, 0.0}
\definecolor{cyan}{rgb}{0.0, 1.0, 1.0}
\definecolor{lightcoral}{rgb}{0.94, 0.5, 0.5}
\definecolor{SlateBlue}{rgb}{0.416, 0.353, 0.804}
\definecolor{LightGreen}{rgb}{0.564, 0.933, 0.564}
\definecolor{Turquoise}{rgb}{0.251, 0.878, 0.816}
\definecolor{BlueGreen}{rgb}{0.643, 0.859, 0.871}
\definecolor{BlueViolet}{rgb}{0.54, 0.17, 0.89}
\definecolor{RAG}{rgb}{0.639, 0.835, 0}
\definecolor{0}{rgb}{0.92, 0.3, 0.26}
\definecolor{1}{rgb}{0.961, 0.851, 0.471}
\definecolor{1_1}{rgb}{1.00, 0.933, 0.698}
\definecolor{2}{rgb}{0.522, 0.784, 0.953}
\definecolor{2_1}{rgb}{0.722, 0.592, 0.871}
\definecolor{2_2_2}{rgb}{0.561, 0.659, 0.843}
\definecolor{3}{rgb}{0.56, 0.93, 0.56}
\definecolor{3_1_1}{rgb}{0.808, 0.996, 0.808}
\definecolor{4}{rgb}{0.988, 0.541, 0.565}
\definecolor{4_1}{rgb}{0.996, 0.796, 0.808}
\definecolor{4_1_1}{rgb}{0.957, 0.847, 0.843}
\definecolor{bounding}{rgb}{0.855, 0.392, 0.357}
\definecolor{6}{rgb}{0.690, 0.612, 0.522}

\definecolor{golden}{RGB}{255, 215, 0}
\definecolor{golden}{RGB}{0, 0, 0}

\definecolor{backbone}{RGB}{255,192,0}
\definecolor{neck}{RGB}{112,173,71}
\definecolor{head}{RGB}{237,125,49}
\definecolor{score}{RGB}{68,72,212}
\definecolor{grade}{RGB}{92,201,207}
\definecolor{rank}{RGB}{0,176,207}

\usepackage{fontawesome}
\usepackage{nicematrix}
\usepackage{calc}

\usepackage{orcidlink}
\usepackage{ragged2e}

\usepackage{tikz}
\usetikzlibrary{calc,positioning}

\newlength{\myMheight}
\usepackage{bbding,pifont}

\usepackage[misc]{ifsym}

\usepackage{times}

\usepackage{etoolbox}
\makeatletter
\patchcmd{\thebibliography}
  {\settowidth}
  {\setlength{\itemsep}{0pt plus 0.3ex}
   \setlength{\parskip}{0pt}
   \settowidth}
  {}{}
\makeatother

\graphicspath{{./figs/}{./figs/imgs/}}
\usepackage{overpic}
\usepackage{subfigure}

\usepackage{subcaption}
\usepackage{xstring}   

\colorlet{highlightcolor}{red!70!black}
\newif\ifMShighlight
\MShighlighttrue      
\MShighlightfalse     

\newif\ifMSinputmode
\MSinputmodefalse

\usepackage[pagewise]{lineno}
\linenumbers

\ifMShighlight
    \else
    
\fi

\makeatletter
\let\MS@originput\input
\renewcommand{\input}[1]{\IfBeginWith{#1}{pr/r2/}{
    \ifMShighlight
      \begingroup
        \MSinputmodetrue        \linelabel{start:#1}\color{highlightcolor}\arrayrulecolor{highlightcolor}\MS@originput{#1}\arrayrulecolor{black}\linelabel{end:#1}\endgroup
      \ignorespaces
    \else 
      \MS@originput{#1}\ignorespaces
    \fi
  }{\MS@originput{#1}\ignorespaces
  }}
\makeatother

\hypersetup{%
  linkcolor={blue},
  urlcolor={magenta},
  citecolor={blue}
}

\def\tsc#1{\csdef{#1}{\textsc{\lowercase{#1}}\xspace}}
\tsc{WGM}
\tsc{QE}
\tsc{EP}
\tsc{PMS}
\tsc{BEC}
\tsc{DE}

\makeatletter
\renewcommand{\vspace}{\@ifstar\@gobble\@gobble}
\makeatother
\usepackage{fancyvrb}
\begin{document}
\let\WriteBookmarks\relax
\def\floatpagepagefraction{1}
\def\textpagefraction{.001}

\shorttitle{PiT: Progressive Diffusion Transformer}

\shortauthors{J. Wu et~al.}

\title [mode = title]{PiT: Progressive Diffusion Transformer}

\author[]{Jiafu Wu}[orcid=0000-0002-1036-5076]
\fnmark[1]

\author[]{Yabiao Wang}[orcid=0000-0002-6592-8411]
\fnmark[1]

\author[]{Jian Li}

\author[]{Jinlong Peng}

\author[]{Yun Cao}

\author[]{Chengjie Wang}

\author[]{Jiangning Zhang}

\author[1]{Yong Liu}
\cormark[1]

\fntext[fn1]{Equal contribution.}
\cortext[cor1]{Corresponding author.}

\affiliation[1]{organization={Institute of Cyber-Systems and Control, Zhejiang University},
    city={Hangzhou},
    postcode={310027}, 
    country={China}}



\begin{abstract}
Diffusion Transformers (DiTs) achieve remarkable performance within image generation. Conventionally, DiTs are constructed by stacking serial isotropic transformers, which face significant quadratic computational cost. However, through empirical analysis, we find that DiTs do not rely as heavily on long-distance information as previously believed. In fact, most layers exhibit significant redundancy in long-distance computation. Additionally, conventional attention mechanisms suffer from low-frequency inertia, limiting their efficiency. To address these issues, we propose \textbf{Pseudo Shifted Window Attention (PSWA)}, which fundamentally mitigates long-distance attention redundancy. PSWA achieves moderate global-local information through \textbf{Static Window Attention}. It further utilizes a high-frequency bridging branch to  enrich the high-frequency information and strengthen inter-window connections. Furthermore, we pioneer the concept of \textbf{Kth-order Attention} and propose the \textbf{Progressive Coverage Channel Allocation (PCCA)} strategy that captures high-order attention by reallocating the existing channel budget. Based on these innovations, we propose a series of Pseudo \textbf{P}rogressive D\textbf{i}ffusion \textbf{T}ransformer (\textbf{PiT}). Extensive experiments show superior performance of PiT; for example, PiT-L achieves 54\% FID improvement over DiT-XL/2 with less computation.
\end{abstract}




\thispagestyle{empty}










\maketitle

\section{Introduction}
\label{sec:intro}
Diffusion Transformer (DiT)~\cite{peebles2023scalable} first utilize the Transformer as the basic architecture of the generation network. The time step and other conditions are injected into the Transformer blocks in the form of scale-shift~\cite{peebles2023scalable} or cross-attention~\cite{chen2021crossvit} to guide the diffusion generation trajectory. However, the introduction of the Transformer into the domain of vision generation by DiT is inelegant. DiT completely discards the U-Net~\cite{ronneberger2015u} architecture that predominantly consists of convolutions in previous works, and instead fully employs the Transformer~\cite{vaswani2017attention} blocks for global modeling. Although scaling up the model achieves state-of-the-art performance, this naturally raises an important question: Are such extensive global computations truly necessary in Diffusion Transformers?

\begin{figure}
\centering
 {
 \includegraphics[width=\linewidth]{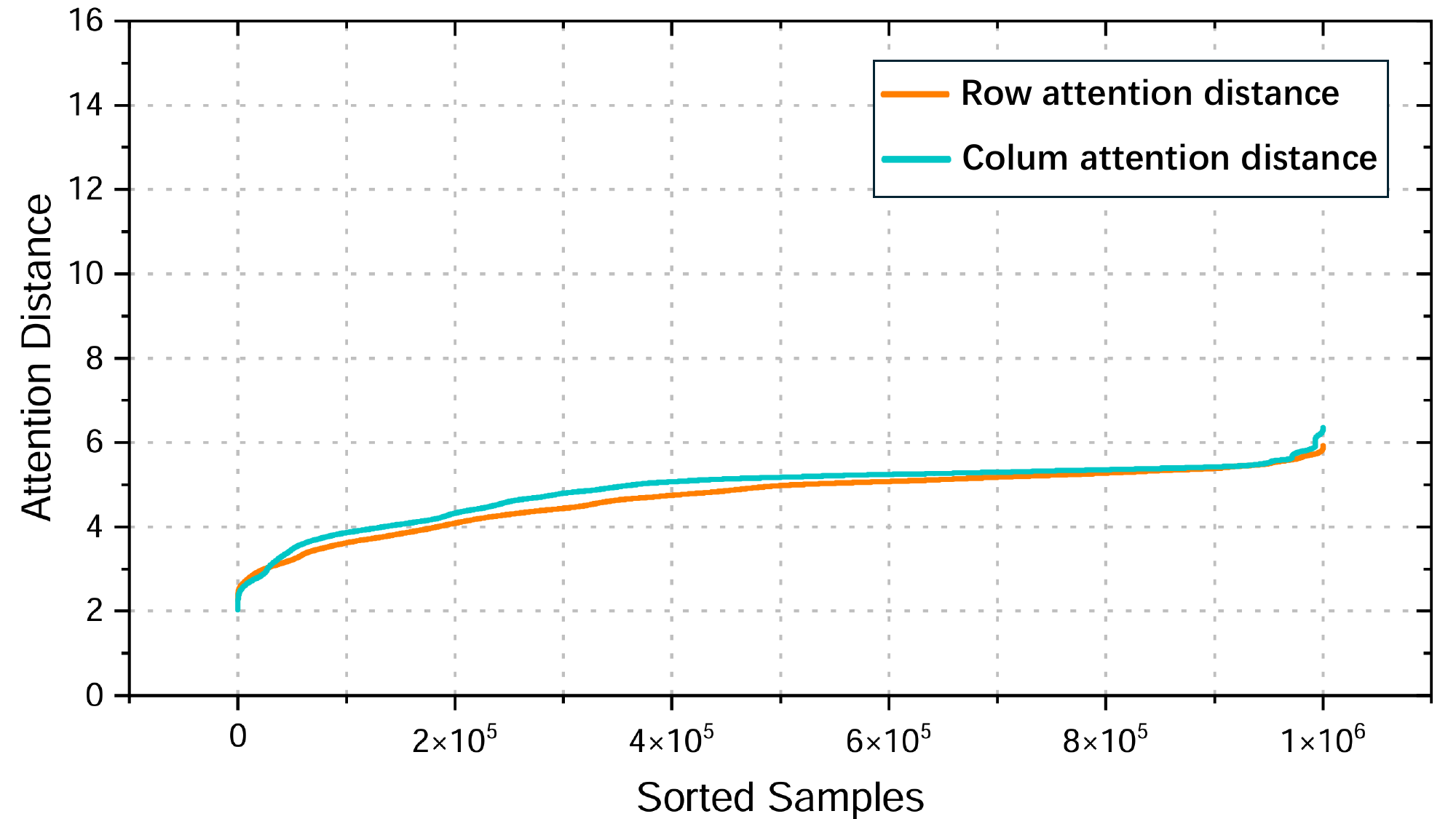}}
	\caption{Row and column attention distance distribution. The x-axis denotes the number of sampled attention maps, and the y-axis denotes the row/column attention distance in token units, where one unit corresponds to the distance between adjacent tokens. We sample one million attention maps to compute row and column attention distances~(Equation~\ref{equ:equ3}) from DiT's attention maps. With a map edge length of 16, 99\% of tokens have row and column attention distances under 6, indicating low long-distance attention demand and high computational redundancy.}
\label{fig2}
\end{figure}

To explore the dependence distance of diffusion full-size attention, we calculate the average tokens' attention distance in each DiT layer. Figure~\ref{fig2} reveals that for a 16×16 latent map, 99\% of tokens attend to neighbors within a row/column distance of 2-6 (see Equation~\ref{equ:equ3} for distance metric). Thus, \textit{\textbf{DiT relies less heavily on long-distance attention than previously assumed}}. Inspired by this finding, we re-evaluate the effect of restricting the Transformer's scope, a strategy that drastically reduces computational overhead. We introduce {\textbf{Static Window Attention}} (Equation~\ref{equ:win_atten}), which constrains tokens to calculate dependencies only within their local window. Our experiments reveal that, even with an identical architecture, the window attention mechanism leads to superior performance, thereby confirming our initial conjecture.

Another challenge is how to establish information connections between different windows. In Swin Transformer~\cite{liu2021swin}, the shifted window approach is a classic method of supplementing window connections. As illustrated in Figure~\ref{fig3} (a), by shifting, the windows create overlapping regions, effectively establishing window connections. However, these methods come with a substantial computational cost, as the model expends excessive computational resources in the overlapping areas.  

Furthermore, as shown in Figure~\ref{fig4}, we observe that the transformer in DiT excessively over-learns low-frequency information in the latent space, while lacking high-frequency details. To compensate for the deficiency of high-frequency information and establish a global information bridge between windows, we propose a novel {\textbf{Pseudo Shifted Window Attention (PSWA)}} (Equation~\ref{equ:pswa1}, \ref{equ:pswa2}). As illustrated in Figure~\ref{fig3}(b) and Figure~\ref{fig1}, each block is divided into a static window attention branch and a high-frequency bridging branch constructed by depthwise separable convolution(DWConv)~\cite{Chol17Xception}. The high-frequency bridging branch not only supplements high-frequency information but also cleverly covers the edge regions of the static windows, constructing a bridge between windows to supplement information integration.

Moreover, we define conventional self-attention as first-order attention (Equation~\ref{equ:kth_attn}), which describes the pairwise similarity between tokens themselves. As shown in Figure~\ref{fig6}, first-order similarity(Equation~\ref{equ:equ11}) tends to bias of low-level feature such as color and texture and fail to capture the intrinsic relationships among tokens. To address it, we propose a novel {\textbf{Progressive Coverage Channel Allocation (PCCA)}} strategy to capture {\textbf{Kth-order attention}} (Equation~\ref{equ:kth_attn}) without additional parameters or FLOPs beyond the base architecture. Our proposed Kth-order attention has a more effective attention mechanism than first-order attention.

In summary, our main contributions are as follows:
\begin{itemize}
    \item We observe that Diffusion Transformers DiT does not rely on long-distance attention modeling as traditionally assumed, and propose a \textbf{Static Window Attention} to mitigate the computational redundancy.
    \item We propose the \textbf{Pseudo Shifted Window Attention (PSWA)}, introducing a high-frequency bridging branch to supplement information connection between windows while alleviating the over-learning of low-frequency information issue.
    \item We pioneer the concept of\textbf{ Kth-order attention} in vision generation and propose a novel \textbf{Progressive Coverage Channel Allocation (PCCA)} strategy to effectively capture Kth-order attention by reallocating the existing channel budget.
    \item Extensive experiments demonstrate that PiTs surpass state-of-the-art models in both isotropic and U-shape architectures. Ablation studies further validate the effectiveness and rationality of our proposed components.
\end{itemize}

\section{Related Work}

\subsection{Vision Transformers}

Compared with convolutional backbones such as ResNet~\cite{he2016deep}, Vision Transformer (ViT)~\cite{dosovitskiy2020image} represents an image as patch tokens and processes them with a stack of Transformer blocks. DeiT~\cite{touvron2021training} improves ViT training when less data is available. PVT~\cite{wang2021pyramid} builds a feature pyramid for dense prediction, while T2T-ViT~\cite{yuan2021tokens} gradually combines nearby tokens to capture local structure. Broadly, these vision models can be categorized into two architectural paradigms: isotropic~\cite{dosovitskiy2020image} and pyramid~\cite{wang2021pyramid}.

Transformers have also been used in many pattern recognition tasks. DETR~\cite{carion2020end} treats object detection as direct set prediction. Mask2Former~\cite{cheng2022masked} uses masked attention for semantic, instance, and panoptic segmentation. IPT~\cite{chen2021pre} applies a pre-trained Transformer to image restoration. Vision-by-Prompt~\cite{wang2025vision} uses visual prompts for composed video retrieval. These works show that Transformer features can support classification, detection, segmentation, restoration, and video understanding.

Several models combine attention with local or multi-scale processing. CvT~\cite{wu2021cvt} adds convolution to token embedding and projection. CoAtNet~\cite{dai2021coatnet} combines convolution and attention at different stages. ViTAE~\cite{xu2021vitae} adds spatial bias to improve local feature learning. EMO~\cite{emo} develops an efficient block for mobile vision models. These designs show that local details and efficient token interaction remain important in Transformer backbones.

Global self-attention captures long-range relations, but its cost grows quickly with the number of tokens. Voita et al.~\cite{voita-etal-2019-analyzing} show that attention heads often learn different roles and that some heads contribute little. Bian et al.~\cite{bian2021attention} also find considerable redundancy in attention computation. Swin Transformer~\cite{liu2021swin} reduces the cost by using shifted local windows. Neighborhood Attention Transformer~\cite{hassani2023neighborhood} lets each token attend to a nearby sliding neighborhood. Twins-SVT~\cite{chu2021twins} alternates local attention and reduced global attention, while CMT~\cite{guo2022cmt} combines convolutional local features with global attention.

\subsection{Diffusion Transformers}

Denoising Diffusion Probabilistic Models~\cite{ho2020denoising} generate samples through repeated denoising with a convolutional U-Net. Latent Diffusion Models~\cite{rombach2022high} perform this process in a compressed latent space to reduce the cost of high-resolution generation. DiT~\cite{peebles2023scalable} replaces the U-Net denoiser with an isotropic Transformer. U-DiT~\cite{tian2024u} introduces token downsampling and a U-shaped structure to reduce computation. All are Worth Words~\cite{bao2023all} further studies how ViT-style backbones should be designed for diffusion models.

Other studies improve diffusion Transformers through training, conditioning, or input design. MDT~\cite{gao2023masked} adds masked latent modeling during diffusion training, and MDTv2~\cite{gao2023mdtv2} improves its masking and training strategy. UniDiffuser~\cite{bao2023one} uses one Transformer to model several image and text distributions. FiT~\cite{lu2024fit} supports different image sizes and aspect ratios. PixArt-$\alpha$~\cite{chen2023pixart} reduces the training cost of large text-to-image Transformers. DiffiT~\cite{hatamizadeh2024diffit} changes attention according to the diffusion timestep. Scaling Rectified Flow Transformers~\cite{pmlr-v235-esser24a} applies a multimodal Transformer and rectified flow to high-resolution text-to-image generation.

\begin{figure*}
    \centering
    \subfigure[Shifted window]{
        \includegraphics[width=0.3\linewidth]{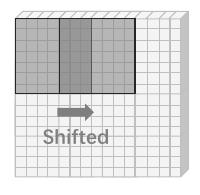}
    }\subfigure[Pseudo shifted (Ours)]{
        \includegraphics[width=0.28\linewidth]{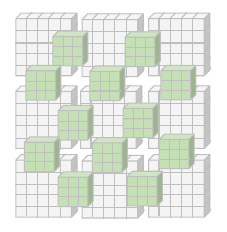}
    }
    \caption{Different window movement mechanisms of Attention. (a) sliding windows to create overlapping regions~\cite{liu2021swin}; (b) parallel high-frequency bridging branches between windows via n×n DWConv~\cite{Chol17Xception}.
    }
\label{fig3}
\end{figure*}

\begin{figure*}
    \centering
    \subfigure[DiT]{
        \includegraphics[width=0.28\linewidth]{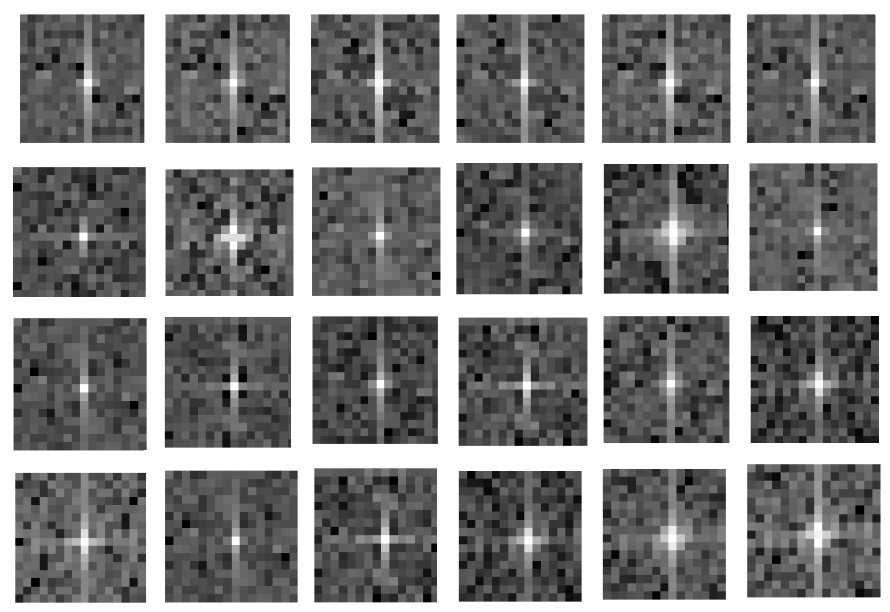}
    }\subfigure[PiT]{
        \includegraphics[width=0.3\linewidth]{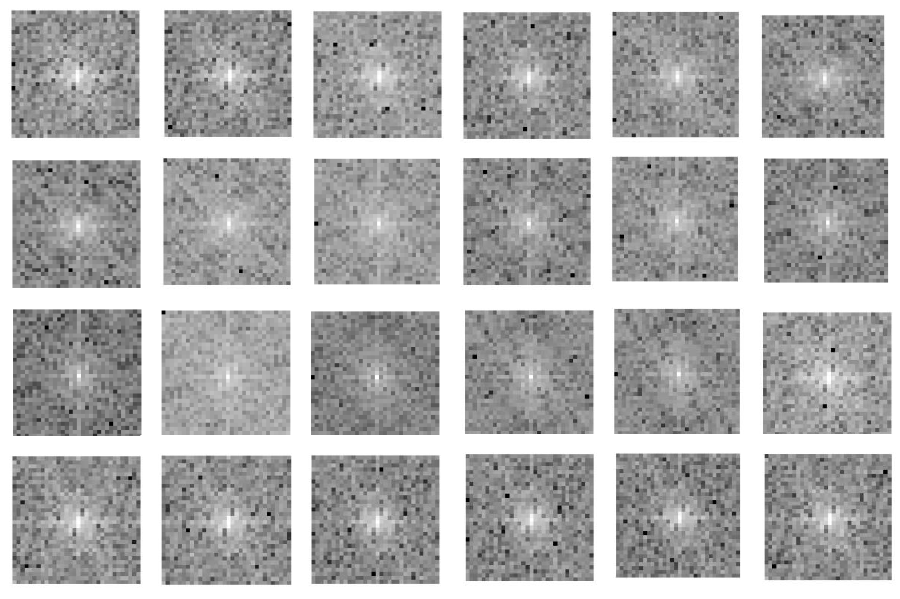}
    }
    \caption{The Fourier transform frequency maps from DiT and PiT across different layers. In each image, the center represents low-frequency components, while the surrounding areas represent high-frequency components. The brightness and darkness variations indicate the magnitude of these components. We additionally annotate the low-frequency and high-frequency energy ratios in the visualization to provide quantitative evidence. Quantitatively, DiT's high-frequency energy ratio is only 24.37\%, while PiT achieves 53.19\%, confirming that PSWA substantially enriches high-frequency representation. Figure (a) illustrates DiT's heavy reliance on low-frequency information, lacking high-frequency details. Figure (b) demonstrates that PSWA enables the model to capture richer high-frequency information while preserving low-frequency structure.}
\label{fig4}
\end{figure*}

Recent work also connects diffusion Transformers with pattern recognition tasks. VisionLLaMA~\cite{chu2025visionllama} studies LLaMA-style components for visual backbones. Token-aware and step-aware acceleration~\cite{zhen2025token} reduces diffusion inference cost by considering token importance and denoising steps. DiffTrajectory~\cite{li2025difftrajectory} applies diffusion modeling to trajectory prediction. VRoPE~\cite{liu-etal-2025-vrope} improves the position encoding of spatiotemporal tokens in video-language models. Wan-Animate~\cite{cheng2025wananimate} uses a video diffusion Transformer for character animation and serves as the baseline in our video experiment.

\section{Method}
\label{sec:exp}
\subsection{Revisiting the Diffusion Transformers}
Defining \(\mathbf{x}_t \in \mathbb{R}^{n \times d}\) as the noisy input at time step \(t\). In the reverse denoising process, \(\mathbf{x}_{t - 1}\) is formulated as:
\begin{align}
\centering
\mathbf{x}_{t-1} = \boldsymbol{\mu}_\theta(\mathbf{x}_t, t) + \boldsymbol{\Sigma}_\theta^{1/2}(\mathbf{x}_t, t)\boldsymbol{\epsilon}, \quad \boldsymbol{\epsilon} \sim \mathcal{N}(\mathbf{0}, \mathbf{I}),
\label{equ:equ1}
\end{align}
where \(\boldsymbol{\mu}_\theta\) and \(\boldsymbol{\Sigma}_\theta\) are the predicted mean and covariance matrix by Diffusion Transformer blocks.

At the microscopic level, for a layer \(l\) with the hidden state \(\mathbf{h}_l \in \mathbb{R}^{n \times d}\), the Diffusion Transformer Block can be expressed as:
\begin{align}
\mathbf{h}_{l}' = \mathbf{h}_l + \mathrm{MHSA}\left(\mathrm{Norm}(\mathbf{h}_l; \mathbf{t})\right),  \\
\mathbf{h}_{l+1} = \mathbf{h}_{l}' + \mathrm{FFN}\left(\mathrm{Norm}(\mathbf{h}_{l}'; \mathbf{t})\right),
    \label{equ:equ2}
\end{align}
where MHSA is the multi-head self attention mechanism. 

\subsection{Exploring the Intrinsic Dependence Distance}
We carefully study the tokens dependence distance in the Multi-Head Self-Attention (MHSA) part of DiT. Specifically, we quantify the mean dependence distance across attention maps during batch generation with a pre-trained DiT-B/2. For a single-head attention map, we formally define the row and column attention distance as:
\begin{equation}
\begin{aligned}
D_{\text{row}}(\mathbf{A}) = \frac{\sum_{i,j} \left| \mathbf{r}_j - \mathbf{r}_i + 1 \right| \cdot A_{ij}}{\sum_{i,j} A_{ij}},  \\
D_{\text{col}}(\mathbf{A}) = \frac{\sum_{i,j} \left| \mathbf{c}_j - \mathbf{c}_i + 1 \right| \cdot A_{ij}}{\sum_{i,j} A_{ij}},
    \label{equ:equ3}
\end{aligned}
\end{equation}
where $A_{ij}$ denotes the similarity \textit{(cosine)} between token \(i\) and token \(j\) in the attention map, and $\mathbf{r}_i$ is the row number of token $i$, $\mathbf{c}_i$ is the colum number of token $i$.

We randomly evaluate the attention distances across one million attention maps. As depicted in Figure \(\ref{fig2}\), when the spatial height and width of the map in the shallow layer are both 16, the majority of rows and columns attention distances fall within the range of 2 to 6, well below the maximum limit of 16. This observation suggests that the long-distance information reliance of DiT is less obvious than previously assumed, and the rapid convergence of DiT is predominantly driven by local information.
\begin{figure*}[]
\centering
 {
 \includegraphics[width=0.7\textwidth]{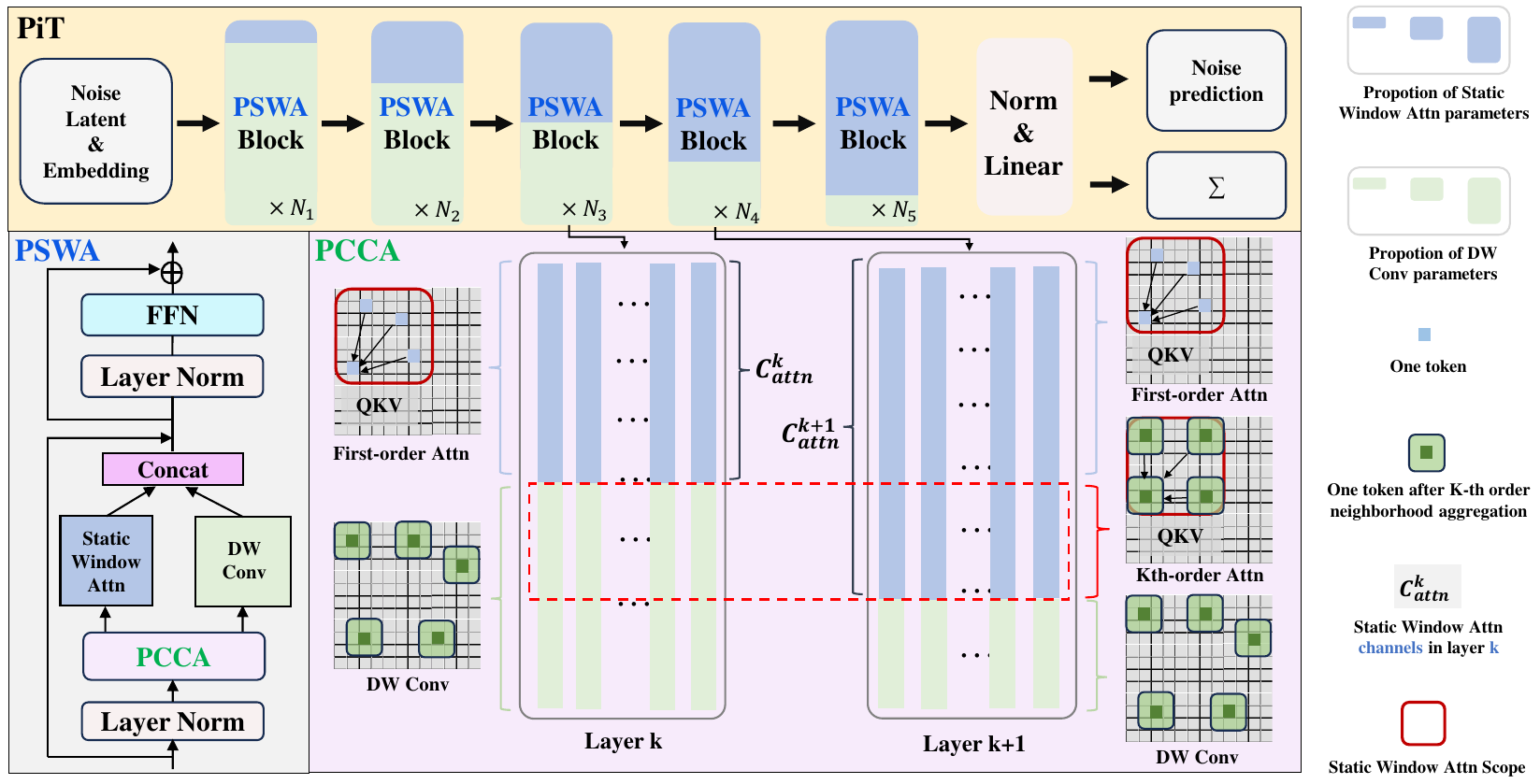}}
    \caption{The overall framework of PiT. The Static Window Attention and DWConv are the core operations of the PSWA blocks. We also employ the Progressive Coverage Channel Allocation (PCCA) strategy to implement Kth-order attention by reallocating the existing channel budget. The detailed definition of Kth-order neighborhood aggregation is provided in Eq.~\ref{equ:equ10}. In PiT, this aggregation is mainly performed by DWConv, which aggregates the neighboring features around each token.}
    \label{fig1}
\end{figure*}

Therefore, we suggest using the {\textbf{Static Window Attention}} (Equation \ref{equ:win_atten}) mechanism in the diffusion framework, which offers a balanced dependency range and computational efficiency.
\begin{align}
\label{equ:win_atten}
\text{WinAttn}(X) =  \bigcup_{w}\left( \frac{Q_w K_w^\top}{\sqrt{d}} + P_w \right) V_w,
\end{align}
where X is the input, $Q_w$, $K_w$, and $V_w$ are the linear transformation outputs of X in the window $w$. $P_w$ is the positional encoding within window \(w\).

\subsection{Pseudo Shifted Window Attention (PSWA)} 
While the Static Window Attention in Equation \ref{equ:win_atten} successfully reduces global computational redundancy, \textit{\textbf{two critical challenges remain}}: (1) inter-window information exchange, and (2) the inherent low-frequency bias in attention mechanisms.

\textbf{Connecting different windows.} It is crucial to establish information connections between different window attention mechanisms. Figure \ref{fig3} (a) illustrate that connections between windows can be supplemented by sliding windows to create overlaps between windows. However, both methods cause the model to consume a large amount of computational resources in the overlapping regions; therefore, they are not the optimal choices.

\textbf{Attention mechanism has a low-frequency bias.} We observe that DiT exhibits a strong bias toward low-frequency information in generative tasks. As illustrated in Figure \(\ref{fig4}\)(a), the DiT feature maps from pure attention show rich low frequencies in the frequency domain, emphasizing macro structures but ignoring detailed and fine-grained information (high-frequency).

A simple DDPM/DDIM interpretation also explains why high-frequency prediction matters. In DDPM, the clean latent predicted from the noisy latent can be written as
\begin{align}
\hat{\mathbf{x}}_0 = \frac{\mathbf{x}_t - \sqrt{1-\bar{\alpha}_t}\,\boldsymbol{\epsilon}_\theta(\mathbf{x}_t,t)}{\sqrt{\bar{\alpha}_t}} .
\label{equ:freq_x0}
\end{align}
Let \(\mathcal{H}(\cdot)\) denote a high-pass operator in the Fourier domain. Since \(\mathbf{x}_t = \sqrt{\bar{\alpha}_t}\mathbf{x}_0 + \sqrt{1-\bar{\alpha}_t}\boldsymbol{\epsilon}\), the high-frequency reconstruction error satisfies
\begin{align}
\mathcal{H}(\hat{\mathbf{x}}_0-\mathbf{x}_0)
= -\sqrt{\frac{1-\bar{\alpha}_t}{\bar{\alpha}_t}}\,\mathcal{H}(\boldsymbol{\epsilon}_\theta-\boldsymbol{\epsilon}).
\label{equ:freq_error}
\end{align}
Therefore, reducing the high-frequency error of the denoiser directly reduces the high-frequency reconstruction error of the generated latent. DDIM uses a deterministic update that is also linear in \(\hat{\mathbf{x}}_0\) and \(\boldsymbol{\epsilon}_\theta\), so the same conclusion applies. This provides a theoretical motivation for preserving high-frequency cues: edges, contours, and textures are mainly represented by high-frequency components, and better recovery of these components improves visual fidelity.

To establish information bridges between windows and supplement high-frequency detail information, we propose a novel Pseudo Shifted Window Attention (\textbf{PSWA}) mechanism. As illustrated in the PSWA part of Figure~\ref{fig1}, the image features are split into two components in the channel space: one is the Static Window Attention branch, and the other is the high-frequency bridging branch via depthwise separable convolution (DWConv). Mathematically, PSWA can be described as follows:
\begin{equation}
\begin{aligned}
\label{equ:pswa1}
\text{PSWA}(X) = FFN(\text{Norm}((f(\text{Norm}\left( X \right)) + X))) + X
\end{aligned}
\end{equation}
\begin{align}
\label{equ:pswa2}
f(X) = \text{Concat}\left( \text{WinAttn}(X_{attn}),\ \text{DWConv}(X_{c}) \right)
\end{align}
where \( X \) is the input, and it is split into two parts at the channel level: \( X_{\text{attn}} \) is used for the Static Window Attention branch, and \( X_c \) is used for the high-frequency bridging branch (DWConv). Norm is the Layer Norm layer, and FFN is the feed-forward network (linear layer).  

\textbf{Complexity comparison.} Let $n$ denote the number of tokens. Standard global attention requires $\mathcal{O}(n^2)$ pairwise attention computation. In PSWA, the dominant attention computation is performed only in the static-window branch. Under an even channel split, the simplified attention cost is $(n/2)\times(n/2)=n^2/4$, plus the linear complexity of the DWConv bridging branch. Therefore, PSWA reduces the dominant quadratic attention cost while using DWConv to maintain local and cross-window information exchange.

Figure~\ref{fig3}(b) better illustrates how PSWA connects different static windows from the pixel perspective. The dense convolution kernels will always have many falling on the adjacent boundaries of different windows (the green part in figure~\ref{fig3}(b)), and they effectively connect different windows. Additionally, Depthwise Convolution (DWConv) is known for its strong high-frequency extraction capabilities. As depicted in Figure \ref{fig4}(b), when PSWA uses DWConv to connect windows, it significantly supplements high-frequency details in the frequency domain. It is also important to clarify the meaning of zero additional cost in PCCA. PCCA itself only changes the channel allocation across layers and introduces no parameters. For the channels assigned to the high-frequency branch, DWConv-based local aggregation replaces the QKV self-attention computation that these channels would otherwise perform in the attention branch. Thus, Kth-order information is captured by reallocating the existing channel budget rather than by adding an extra branch on top of full-channel attention.

\subsection{Progressive Coverage Channel Allocation (PCCA)}
\label{sec3.4}
Before introducing Progressive Coverage Channel Allocation (PCCA), we first need to present three important new concepts: Kth-order neighborhood, Kth-order similarity, and Kth-order attention.

\textbf{Kth-order neighborhood.} Considering an input sequence represented as a set of points. For any position \(i\) in this sequence, we define its Kth-order neighborhood \(\mathcal{N}_K(i)\) as the set of positions \(j\) that satisfy the spatial condition:
\begin{align}
\mathcal{N}_K(i) \triangleq \left\{ j \ \middle| \  \max\left(|x_j - x_i|, |y_j - y_i|\right) \leq K - 1 \right\}
\label{equ:equ10}
\end{align}
where \(x\) and \(y\) are the coordinate components of positions \(i\) and \(j\). To put it simply, the Kth-order neighborhood of a central node `i' comprises the nodes within the (K-1) surrounding layers.

\textbf{Kth-order similarity.} Assuming an aggregation function \(\Phi\) that processes the Kth-order neighborhood to one feature representation. The Kth-order similarity between positions \(i\) and \(j\) is then defined as:
\begin{align}
S_K(i, j) \triangleq \left\langle \Phi(\mathcal{N}_K(i)), \Phi(\mathcal{N}_K(j)) \right\rangle,
\label{equ:equ11}
\end{align}
where \(\langle \cdot, \cdot \rangle\) is the inner product operation, and \(\Phi\) is a \textbf{{Kth-order neighborhood aggregation}} (e.g., DWConv). If \(K=1\), the Kth-order similarity degenerates to the \textbf{first-order similarity}.

\textbf{Kth-order attention.} Kth-order attention is an attention mechanism that calculates the attention map using Kth-order similarity, and its mathematical definition is as follows:
\begin{align}
\label{equ:kth_attn}
\text{$Attn_{K}$}(X) =  \left( S_K + P \right) V.
\end{align}
where $X$ is the input, $S_K$ is the similarity map computed by the Kth-order similarity in Eq.~\ref{equ:equ11}, $P$ is the positional encoding, and $V$ is the output of $X$ after a linear layer. If \(K=1\), the Kth-order attention degenerates to \textbf{first-order attention}.

\begin{figure*}
    \centering
    \subfigure[DiT]{
        \includegraphics[width=0.49\textwidth]{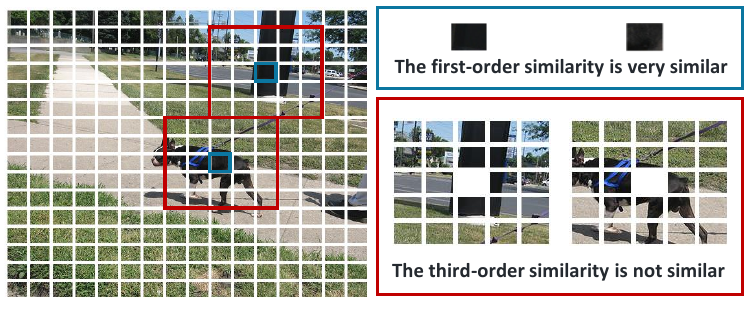}
    }\subfigure[PiT]{
        \includegraphics[width=0.49\textwidth]{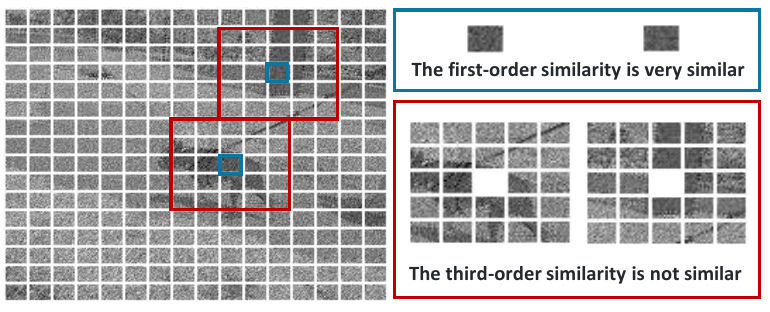}
    }
    \caption{The first-order similarity measure tends to prioritize low-level features such as color and texture. As illustrated by the blue box, two black image patches appear similar in their low-level features, even though one represents a dog's fur and the other a roadside post. The first-order similarity measure thus leads to a strong correlation between them, potentially misleading the model. In contrast, when higher-order similarity is considered, as shown in the red box, the expanded receptive field captures the semantic differences between these patches.}
\label{fig6}
\end{figure*}

The first-order similarity only measures the similarity of the tokens themselves, not considering the similarity of their neighboring tokens. The first-order similarity may be inadequate for capturing the intricate, high-level relationships between tokens. As shown in Figure~\ref{fig6}, this measure is biased towards low-level features such as color and texture. In Figure~\ref{fig6}, the two black nodes: one is a part of the dog's fur, and one is a part of the street lamp. They show extremely high similarity both in pixel space and latent space. However, they belong to completely different categories and should not exhibit such a strong similarity dependency. It means that two tokens with similar color and texture but no semantic relationship will exhibit a strong correlation, while tokens with meaningful semantic connections but differing low-level features are often disregarded in the diffusion transformer layers. This limitation compromises the semantic consistency of the generated images or videos.

In contrast, the Kth-order similarity is more semantic, as it doesn't merely describe the similarity of the tokens themselves but rather measures the feature similarity of the surrounding neighbors of the central token. The Kth-order similarity can effectively compensate for the shortcomings of the first-order similarity. It encourages attention connections where both the tokens themselves and their neighbors are similar, and simultaneously penalizes the misleading attention connections where the tokens are similar to each other but dissimilar to their surrounding neighbors. For example, as shown in Figure~\ref{fig6}, when we consider neighbors within two circles around the black nodes, the difference in features becomes apparent. In other words, the third-order similarity between these two black nodes is very low.  

\textbf{Capturing Kth-order similarity.} The higher-order similarity with a larger similarity receptive field needs more complex calculations. An intuitive way to capture it is to design a special module to additionally calculate the Kth-order similarity, but this will introduce an extra computational burden to the model. 

To capture Kth-order similarity by reallocating the existing channel budget without introducing additional parameters or FLOPs beyond the base architecture, we ingeniously introduce a Progressive Coverage Channel Allocation (\textbf{PCCA}) strategy based on the PSWA. As illustrated in PCCA part of Figure~\ref{fig1}, the tokens of layer k are split into two branches at the channel space. The blue part is the static window attention, and it occupies $C^{k}_{attn}$ channels. \textit{\textbf{The conventional attention within the window is equal to first-order attention}}, as it captures the similarity of the tokens themselves within the window. The other part of the channels is the high-frequency bridging branch. It is implemented by DWConv and mainly captures local and high-frequency information.

\textbf{\textit{The key operation of PCCA is as follows: As the model layer goes deeper, the channel proportion of the static window attention branch increases, and the channel proportion of the high-frequency bridging branch decreases.}}

In the shallow layers of the model, the channel proportion of the high-frequency bridging branch far exceeds that in the static window attention branch. As the model progresses deeper, the proportion of high-frequency bridging branch channels gradually decreases, and part of its channels are transferred to the static window attention branch. As illustrated in the PCCA part of Figure~\ref{fig1}, in layer $k+1$, the channels within the red dashed box are assigned to the static window attention branch. The tokens in these channels have already absorbed sufficient local information from their surrounding neighborhood in layer $k$. Taking the high-frequency bridging branch with a $5 \times 5$ DWConv as an example, the channels in the red area have already captured features from the third-order neighborhood in layer $k$. 
These tokens now embody features from their third-order neighborhood. Consequently, when the static window attention in layer $k+1$ operates on them, it effectively computes a third-order similarity, despite its inherent first-order design. This is expressed mathematically as:
\begin{align}
    S_{ij}^{3} = \mathop \sum \limits_c^{{C^{k+1}_{attn}} - {C^{k}_{attn}}} \left( { {Conv_{5}}(X_i^c) \cdot {Conv_{5}}(X_j^c)} \right),
    \label{equ:equ12}
\end{align}
where $C^{k}_{attn}$ denotes the channels of window attention in the layer $k$, while $C^{k+1}_{attn}$ denotes the channels of window attention in the layer $(k+1)$. $Conv_{5}$ is equal to the aggregation function $\Phi$ in Equation~\ref{equ:equ11}. ${Conv_{5}}(X_i^c)$ is the output of $5\times5$ convolution at token $i$, channel $c$; and ${Conv_{5}}(X_j^c)$ is the output of $5\times5$ convolution at token $j$, channel $c$.

\subsection{Architecture and Implementation Details}
\label{sec:arch_details}
To improve reproducibility, we provide the complete architectural specifications of PiT. PiT adopts a U-shaped multi-scale architecture with five stages: two downsampling stages, one bottleneck stage, and two upsampling stages, linked by skip connections at matching resolutions. All PiT variants use stage depths of [2, 5, 8, 5, 2]. Downsampling is implemented by Conv2d followed by PixelUnshuffle, and upsampling is implemented by Conv2d followed by PixelShuffle. Each PiT block consists of two sub-blocks, each containing adaLN modulation, PSWA, and a convolutional FFN.

\begin{table*}[!htbp]
\centering
\caption{Architecture specifications of PiT variants. $C$ denotes the stage hidden dimension, $C_{attn}$ denotes the channel dimension assigned to the static window attention branch, and the remaining channels are assigned to the high-frequency bridging branch.}
\resizebox{0.8\textwidth}{!}{%
\begin{tabular}{@{}lccccc@{}}
\toprule
\textbf{Model} & \textbf{Depths} & \textbf{$C$ per stage} & \textbf{$C_{attn}$ per stage} & \textbf{Heads} & \textbf{Win / DWConv / MLP} \\
\midrule
PiT-S & [2,5,8,5,2] & [192,224,384,224,224] & [48,80,192,160,192] & 4 & 8 / $3\!\times\!3$ / 2 \\
PiT-B & [2,5,8,5,2] & [240,432,816,432,432] & [80,160,416,320,384] & 8 & 8 / $9\!\times\!9$ / 2 \\
PiT-L & [2,5,8,5,2] & [480,864,1600,864,864] & [120,320,800,560,800] & 10 & 8 / $9\!\times\!9$ / 2 \\
\bottomrule
\end{tabular}%
}
\label{tab:arch_spec}
\end{table*}

Taking PiT-B as an example, the attention channel ratio progressively evolves across the five stages as 33\%, 37\%, 51\%, 74\%, and 89\%. This progressive increase is the core of PCCA: channels processed by DWConv in shallower stages are gradually absorbed into the attention branch in deeper stages, enabling higher-order similarity computation by reallocating the existing channel budget rather than introducing additional parameters or FLOPs beyond the base architecture.

\begin{table}[!htbp]
\centering
\caption{Per-stage channel allocation of PiT-B.}
\resizebox{\linewidth}{!}{
\begin{tabular}{@{}lccccc@{}}
\toprule
Stage & Down-1 & Down-2 & Bottleneck & Up-1 & Up-2 \\
\midrule
Total $C$ & 240 & 432 & 816 & 432 & 432 \\
$C_{attn}$ & 80 & 160 & 416 & 320 & 384 \\
$C_{local}$ & 160 & 272 & 400 & 112 & 48 \\
Attention ratio & 40\% & 37\% & 51\% & 74\% & 89\% \\
\bottomrule
\end{tabular}
}
\label{tab:pcca_stage}
\end{table}

The detailed forward process of one PSWA block is summarized in Algorithm~\ref{alg:pswa}. In practice, we apply 2D RoPE to the window attention branch and use cosine attention with a learnable temperature. The high-frequency bridging branch is implemented by DWConv followed by residual addition, and the two branches are fused by a $1\times1$ convolution.

\begin{algorithm}[t]
\caption{PSWA Block Forward}
\label{alg:pswa}
\begin{algorithmic}[1]
\STATE \textbf{Input:} Feature map $X \in \mathbb{R}^{B \times C \times H \times W}$, condition $c$
\STATE \textbf{Hyperparameters:} $C_{attn}$, $C_{local}=C-C_{attn}$, window size $W_s$
\STATE $\gamma_{msa}, \beta_{msa}, \alpha_{msa}, \gamma_{mlp}, \beta_{mlp}, \alpha_{mlp} \leftarrow \mathrm{adaLN}(c)$
\STATE $X_{norm} \leftarrow \mathrm{LayerNorm2d}(X) \cdot (1+\gamma_{msa}) + \beta_{msa}$
\STATE $X_{attn} \leftarrow X_{norm}[:, :C_{attn}]$ \hfill $\triangleright$ Window attention branch
\STATE $X_{local} \leftarrow X_{norm}[:, C_{attn}:]$ \hfill $\triangleright$ High-frequency bridging branch
\STATE Partition $X_{attn}$ into $W_s \times W_s$ windows
\STATE Compute $Q,K,V$ by linear projection and apply 2D RoPE to $Q,K$
\STATE $Y_{attn} \leftarrow \mathrm{Softmax}(\cos(Q,K)\cdot \exp(\tau))\cdot V$
\STATE $Y_{local} \leftarrow \mathrm{DWConv}_{k\times k}(X_{local}) + X_{local}$
\STATE $Y \leftarrow \mathrm{Conv}_{1\times1}(\mathrm{Concat}(Y_{attn},Y_{local}))$
\STATE $X \leftarrow X + \alpha_{msa}\cdot Y$
\STATE $X \leftarrow X + \alpha_{mlp}\cdot \mathrm{FFN}(\mathrm{LayerNorm2d}(X)\cdot(1+\gamma_{mlp})+\beta_{mlp})$
\STATE \textbf{Output:} $X$
\end{algorithmic}
\end{algorithm}

\section{Experiments}
Based on the above investigations, we design a series of isotropic and U-shaped PiTs. To demonstrate the performance and efficiency of PiTs, we conduct experiments following the training settings of DiT. The state-of-the-art methods involved in the experiment are as follows: DiT~\cite{peebles2023scalable}, MDT~\cite{gao2023masked}, MDTv2~\cite{gao2023mdtv2}, U-DiT~\cite{tian2024u}, DiffT~\cite{hatamizadeh2024diffit}, and PixArt-$\alpha$~\cite{chen2023pixart}.

\subsection{Training and Evaluation Protocol}
\label{sec:training_eval}
We follow the standard IDDPM training protocol. The noise schedule is linear with 1000 diffusion steps, and timesteps are uniformly sampled from $\{0,1,\ldots,999\}$. The model predicts noise $\epsilon$ and learns a variance range interpolating between fixed-small and fixed-large variances. The training objective combines the MSE loss on predicted noise with the KL term for learned variance. We maintain an EMA of model weights with decay 0.9999, and all reported metrics are computed using the EMA model. For class conditioning, we use class-label embeddings with a 10\% dropout probability for classifier-free guidance training. All models are trained in the latent space of sd-vae-ft-ema with scale factor 0.18215, producing $32\times32$ latent maps for $256\times256$ images. Unless otherwise specified, we use AdamW with learning rate $1\times10^{-4}$, no weight decay, global batch size 256, and FP16 mixed precision on 8 NVIDIA A100 80GB GPUs.

For evaluation, we follow the ADM-style evaluation pipeline. We generate 50,000 images using DDPM sampling with 250 denoising steps. Class labels are uniformly drawn from all 1000 ImageNet classes, and no classifier-free guidance is used during metric computation. The reference statistics are computed from the full ImageNet-1K training set using the pre-trained InceptionV3 model from OpenAI's ADM toolkit. All compared models use the same resolution, sampling steps, number of generated samples, reference statistics, and evaluation code. To quantify the robustness of key metrics, for the three U-DiT/PiT scale pairs in Table~\ref{tab2}, the reproduced U-DiT results and our PiT results are reported in mean $\pm$ standard deviation format.

The evaluation metrics are defined as follows. Let $(\mu_r,\Sigma_r)$ and $(\mu_g,\Sigma_g)$ denote the mean and covariance of real and generated features, respectively. FID is computed with InceptionV3 pool3 features:
\begin{equation}
\mathrm{FID}=\|\mu_r-\mu_g\|_2^2+\mathrm{Tr}(\Sigma_r+\Sigma_g-2(\Sigma_r\Sigma_g)^{1/2}).
\end{equation}
sFID follows the same formula but uses spatial Inception features. Inception Score (IS) is defined as
\begin{equation}
\mathrm{IS}=\exp\left(\mathbb{E}_{x}\left[D_{\mathrm{KL}}(p(y|x)\|p(y))\right]\right).
\end{equation}
Precision and Recall follow the $k$-NN manifold protocol. Given the real-image manifold $\mathcal{M}_r$ and generated-image manifold $\mathcal{M}_g$, they are computed as
\begin{equation}
\mathrm{Prec.}
=\frac{1}{|G|}
\sum_{g\in G}
\mathbb{1}[g\in\mathcal{M}_r],
\quad
\mathrm{Rec.}
=\frac{1}{|R|}
\sum_{r\in R}
\mathbb{1}[r\in\mathcal{M}_g].
\end{equation}
Percentage improvements are computed under the same evaluation protocol. For FID, the relative improvement is $(\mathrm{FID}_{baseline}-\mathrm{FID}_{ours})/\mathrm{FID}_{baseline}\times100\%$.

\paragraph{Complexity comparison.}
Let $n$ denote the number of tokens. Standard global attention computes a full $n\times n$ attention map, with complexity $\mathcal{O}(n^2)$. In PSWA, the features are split into a static-window attention branch and a DWConv bridging branch. Under an even split, the dominant attention computation can be simplified as $(n/2)\times(n/2)=n^2/4$, plus the linear complexity of the DWConv branch. This reduction of the quadratic attention term explains the throughput advantage in Table~\ref{tab6}.

\begin{table}[!htbp]
\centering
\caption{Comparison of model parameters and FLOPs.}
\resizebox{\linewidth}{!}{
\begin{tabular}{@{}l|rr|rr|rr@{}}
\toprule
\textbf{Model}    & U-DiT-S   & PiT-S  & U-DiT-B   & PiT-B  & U-DiT-L   & PiT-L   \\
\midrule
\textbf{\#Param.(M)}     &58.90       &58.06         &231.25       &236.07      &916.34       &914.99            \\
\textbf{FLOPs(G)}    &7.15       &7.08         &26.55       &25.56         &102.11       &100.80         \\
\bottomrule
\end{tabular}
}
\label{tab1}
\end{table}

\subsection{Scaling the PiTs Up}
We construct PiTs based on the core operation of PSWA, and scale up the models to compare them with larger-sized DiTs and some previous state-of-the-art (SOTA) models. For a fair comparison, the macro-architecture of PiT adopts a U-shaped~\cite{tian2024u}, and the comparison of parameters and FLOPs with U-DiTs is shown in Table~\ref{tab1}. 
\begin{table}[]
\centering
\caption{The comparison of various PiT sizes with state-of-the-art methods. The resolution is 256$\times$256.}
\resizebox{\linewidth}{!}{
\begin{tabular}{@{}lrrrrrr@{}}
\toprule
\textbf{Model}       & \textbf{FLOPs(G)} & \textbf{FID} $\downarrow$ & \textbf{sFID} $\downarrow$ & \textbf{IS} $\uparrow$ & \textbf{Prec.} $\uparrow$ & \textbf{Rec.} $\uparrow$ \\
\midrule
DiT-S/2          & 5.46          & 68.40                     & -            & -             & -              & -                   \\
DiT-S/2*        & 5.46          & 67.87                     & 9.63            & 20.16             & 0.350              & 0.536                   \\
MDT-S/2     & 6.06         & 53.46               & -                        & -                  & -                    & -                   \\
MDTv2-S/2    & 6.06         & 39.50               & -                        & -                  & -                    & -                   \\
U-DiT-S      & 7.15           & 31.51                     & 8.97                        & 51.62                  & 0.543                    & \textbf{0.633}  \\

\midrule
PiT-S (Ours)      & 7.08          & \textbf{29.89}                     & \textbf{8.78}                        & \textbf{53.28}                  & \textbf{0.550}                    & 0.6231                   \\
\midrule
DiT-B/2           & 21.82          & 43.47                     & -            & -             & -              & -                   \\
DiT-B/2*            & 21.82          & 42.62                     & 8.57            & 33.31             & 0.480              & 0.626                   \\
MDT-B/2       & 23.01         & 34.33               & -                        & -                  & -                    & -                   \\
MDTv2-B/2   & 23.02         & 19.55               & -                        & -                  & -                    & -                   \\
U-DiT-B       & 26.55           & 16.64                     & 6.33                        & 85.15                  & 0.642                    & 0.639  \\

\midrule
PiT-B (Ours)      & 25.56          & \textbf{15.56}                     & \textbf{6.00}                        & \textbf{88.86}                  & \textbf{0.648}                    & \textbf{0.647}                   \\
\midrule
U-ViT-L             & 76.40          & 21.22                     & 6.10                        & 67.64                  & 0.615                    & 0.633                   \\
DiffTi-XL/2        & 118.50          & 36.86                     & 6.53                        & 35.39                  & 0.540                    & 0.613                   \\
PixArt-$\alpha$-XL/2   & 118.40         & 24.75                     & 6.08                        & 52.24                  & 0.612                    & 0.613                   \\
DiT-XL/2            & 114.46          & 19.47                     & -                        & -                  & -                    & -                   \\
DiT-XL/2*           & 114.46          & 19.86                     & 6.19                        & 66.97                  & 0.634                    & 0.622                   \\
MDT-XL/2         & 118.70          & 16.42               & -                        & -                  & -                    & -                   \\
U-DiT-L     & 102.11           & 10.08                     & 5.21                        & 112.44                 & 0.702                    & \textbf{0.631}                   \\
\cmidrule(r){1-7}
PiT-L (Ours)      & 100.80           & \textbf{9.18}                     & \textbf{4.87}                        & \textbf{117.76}                 & \textbf{0.714}                    &  0.620                   \\
\bottomrule
\end{tabular}
}
\label{tab2}
\end{table}
All the models are trained for 400K iterations under the same parameter settings. Table~\ref{tab2} presents the results on the standard ImageNet with a size of 256 $\times$ 256, while Table~\ref{512} reports the results at a resolution of 512 $\times$ 512. In Table~\ref{tab2}, it can be observed that PiT-B achieves an FID metric that is 24.1\% better than that of DiT-XL/2 while having only 23.2\% of the FLOPs of DiT-XL/2. PiT-B also shows an FID improvement of 6.5\% compared to U-DiT-B. Additionally, PiT-L outperforms DiT-XL/2 by 10.68 in the FID metric with fewer FLOPs, which is almost twice the performance of DiT-XL/2. In contrast to the SOTA model U-DiT-L, PiT-L maintains 8.9\% lead. These comparisons cover different model scales and resolutions, showing that the effectiveness of PiT is not limited to the DiT-B/2 setting used in the initial attention-distance motivation.
\begin{table}[!htbp]
\centering
\caption{Comparing PiT against DiT and U-DiT on ImageNet 512 $\times$ 512 generation. Results marked with $*$ denote reproduced results.}
\resizebox{\linewidth}{!}{
\begin{tabular}{@{}lrrrrrr@{}}
\toprule
\textbf{Model 512 $\times$ 512}       & \textbf{Steps} & \textbf{FID} $\downarrow$ & \textbf{sFID} $\downarrow$ & \textbf{IS} $\uparrow$ & \textbf{Prec.} $\uparrow$ & \textbf{Rec.} $\uparrow$ \\
\midrule
DiT-XL/2*          & 400K             & 21.07                      & 6.80                          & 66.14                     & 0.741                       & 0.575                      \\
U-DiT-B*           & 400K             & 15.62                     & 6.82                        & \textbf{92.43}                 & 0.751                   & 0.601                   \\
\midrule
PiT-B (Ours)            & 400K            & \textbf{15.08}                     & \textbf{6.44}                        & 90.04                 & \textbf{0.753}                    & \textbf{0.608}            \\
\bottomrule
\end{tabular}
}
\label{512}
\end{table}

For a more detailed comparison, we carry out a step-by-step comparison between PiTs of both Base and Large sizes and the previous state-of-the-art (SOTA) methods. We respectively report their metrics at 200k, 400K, 600K, 800K, and 1M training iterations. As can be seen from Table~\ref{tab4}, the PiT-L with a smaller model size surpasses the metrics of DiT-XL/2 at 7M iterations when it reaches 1M iterations. The FID of PiT-L is 27.7\% better, while the training cost is reduced by a factor of seven. When compared with U-DiT, PiT-B and PiT-L outperforms U-DiT-B and U-DiT-L at all iterations.

\begin{table}[!htbp]
\centering
\caption{We performed a comprehensive evaluation of PiT models at various training iterations on the ImageNet-1K dataset with a resolution of 256x256. Across all iterations, our PiT models consistently achieved superior FID metrics compared to both DiT and U-DiT.}
\resizebox{\linewidth}{!}{%
\setlength{\extrarowheight}{0pt}%
\renewcommand{\arraystretch}{0.85}%
\footnotesize
\begin{tabular}{@{}lrrrrrr@{}}
\toprule
\textbf{Model}       & \textbf{Steps} & \textbf{FID} $\downarrow$ & \textbf{sFID} $\downarrow$ & \textbf{IS} $\uparrow$ & \textbf{Prec.} $\uparrow$ & \textbf{Rec.} $\uparrow$ \\
\midrule
DiT-XL/2         & 7M             & 9.62                      & --                          & --                     & --                       & --                      \\
\cmidrule(r){1-7}
U-DiT-B            & 200K             & 23.23                     & 6.84                        & 64.42                 & 0.610                    & 0.621                   \\
U-DiT-B~            & 400K             & 16.64                     & 6.33                        & 85.15                 & 0.642                    & 0.639                   \\
U-DiT-B            & 600K             & 14.51                     & 6.30                        & 94.56                 & 0.652                    & 0.643                   \\
U-DiT-B           & 800K             & 13.53                     & 6.27                        & 98.99                 & 0.654                    & 0.645                   \\
U-DiT-B           & 1M             & 12.87                     & 6.33                        & 103.79                 & 0.661                    & 0.653                   \\
PiT-B (Ours)            & 200K    &21.80  &6.34  &67.80  &0.625  &0.626      \\
PiT-B (Ours)            & 400K           & 15.56                     & 6.00                        & 88.86                  & 0.648                    & 0.647                   \\
PiT-B (Ours)            & 600K           & 13.53                     & 6.00                        & 98.04                  & 0.657                    & 0.647                   \\
PiT-B (Ours)            & 800K           & 12.62                     & \textbf{5.96}                        & 103.61                  & 0.661                    & \textbf{0.654}                   \\
PiT-B (Ours)            & 1M             & \textbf{11.83}                     & 5.98                        & \textbf{108.33}                 & \textbf{0.663}                    & 0.650            \\
\cmidrule(r){1-7}
U-DiT-L           & 200K             & 15.26                      & 5.60                        & 86.01                 & 0.685                    & 0.615                   \\
U-DiT-L           & 400K             & 10.08                      & 5.21                        & 112.44                 & 0.702                    & 0.631                   \\
U-DiT-L           & 600K             & 8.71                      & 5.17                        & 122.45                 & 0.705                    & 0.645                   \\
U-DiT-L           & 800K             & 7.96                      & 5.21                        & 131.35                 & 0.705                    & 0.648                   \\
U-DiT-L           & 1M             & 7.54                      & 5.27                        & 135.49                 & 0.706                    & 0.659                   \\
PiT-L (Ours)            & 200K           & 13.76                     & 5.09                        & 90.49                  & 0.693                    & 0.612                   \\
PiT-L (Ours)            & 400K            & 9.18                     & \textbf{4.87}                        & 117.76              & 0.714                  & 0.620                   \\
PiT-L (Ours)            & 600K           & 7.91                      & 4.91                        & 128.24                 & \textbf{0.718}                    & 0.635              \\
PiT-L (Ours)            & 800K           & 7.25                      & \textbf{4.87}                        & 134.15                 & 0.716                    & 0.641                   \\
PiT-L (Ours)            & 1M             & \textbf{6.96}                      & 5.02                        & \textbf{138.84}                 & 0.714                    & \textbf{0.646}          \\
\bottomrule
\end{tabular}}
\label{tab4}
\end{table}

To purely demonstrate the remarkable performance of the Pseudo Shifted Window Attention (PSWA), we conduct comparative experiments on the isotropic architecture of DiT by merely replacing the multi-head self-attention mechanism with PSWA, while keeping all the other structures and hyperparameters unchanged. As can be seen from Table~\ref{tab5}, simply substituting the Multi-Head Self-Attention (MHSA) has achieved substantial improvements across models of different sizes. It is particularly worth noting that the computational cost of PSWA is lower compared to that of MHSA. Table~\ref{tab6} presents a comparison of the inference throughput per second between PiT and DiT across three model sizes. PiT achieves significantly faster inference speeds than DiT for all model sizes.

\begin{table}[!htbp]
\centering
\caption{The comparative experimental results of replacing Multi-Head Self-Attention (MHSA) with Pseudo Shifted Window Attention (PSWA) in DiTs. Results marked with $*$ denote reproduced DiT baselines; unmarked DiT baseline rows are taken from the original DiT paper. PiT rows are our PSWA replacement results.}
\resizebox{\linewidth}{!}{
\begin{tabular}{@{}lrrrrrr@{}}
\toprule
\textbf{Model}       & \textbf{FLOPs(G)} & \textbf{FID} $\downarrow$ & \textbf{sFID} $\downarrow$ & \textbf{IS} $\uparrow$ & \textbf{Prec.} $\uparrow$ & \textbf{Rec.} $\uparrow$ \\
\midrule
DiT-S/2             & 5.46           & 68.40                     & --                          & --                     & --                       & --                      \\
DiT-S/2*            & 5.46           & 67.40                     & 11.93                       & 20.44                  & 0.368                    & 0.559                   \\
CvT-S/2* & 5.41 & 80.32 & 15.04 & 18.17 & 0.322 & 0.516 \\
PiT-S/2 (Ours)      & 5.30           & \textbf{61.24}                    & \textbf{10.37}                        & \textbf{23.02}                  & \textbf{0.380}                    & \textbf{0.576}                   \\
\cmidrule(r){1-7}
DiT-B/2             & 21.82          & 43.47                     & --                          & --                     & --                       & --                      \\
DiT-B/2*           & 21.82          & 42.84                     & 8.24                        & 33.66                  & 0.491                    & 0.629                   \\
CvT-B/2* & 20.84 & 47.36 & 11.49 & 27.92 & 0.397 & 0.574 \\
PiT-B/2 (Ours)      & 20.13          & \textbf{39.19}                     & \textbf{7.67}                        & \textbf{37.76}                  & \textbf{0.500}                    & \textbf{0.635}                   \\
\cmidrule(r){1-7}
DiT-L/2            & 77.53          & 23.33                     & --                          & --                     & --                       & --                      \\
DiT-L/2*           & 77.53          & 23.27                     & 6.35                        & 59.63                  & 0.611                    & 0.635                   \\
CvT-L/2* & 76.64 & 27.03 & 8.75 & 51.38 & 0.552 & 0.597 \\
PiT-L/2 (Ours)      & 72.30          & \textbf{21.25}                     & \textbf{5.72}                        & \textbf{65.41}                 & \textbf{0.724}                    & \textbf{0.638}                   \\
\bottomrule
\end{tabular}
}
\label{tab5}
\end{table}

\begin{table}[!htbp]
\centering
\caption{Comparison of FP32 training throughput between DiTs and PiTs on the same GPUs.}
\begin{tabular}{@{}l|rrr@{}}
\toprule
\multirow{4}{*}{\shortstack{Throughput$\uparrow$\\ (imgs/s)}} & DiT-S/2   & DiT-B/2   & DiT-L/2  \\
    & 497        & 272       & 104          \\
\cmidrule{2-4}
   & PiT-S/2  & PiT-B/2   & PiT-L/2   \\
   & \textbf{1225}        & \textbf{618}      & \textbf{248}            \\
\bottomrule
\end{tabular}
\label{tab6}
\end{table}

\subsection{Ablation Study}
The default static window size $W_s$ is set to 8 for all PiT variants, as specified in Table~\ref{tab:arch_spec}. To systematically analyze the sensitivity of generation quality to this hyperparameter, we conduct a window-size ablation on PiT-S trained for 400K iterations on ImageNet-1K. As shown in Table~\ref{tab_winsize}, we evaluate window sizes $W_s \in \{4, 8, 16\}$. The results demonstrate that $W_s = 8$ achieves the best overall performance, confirming our default design choice.

\begin{table}[!htbp]
\centering
\caption{Ablation study on the static window size $W_s$ of PiT-S. All results were trained on ImageNet-1K for 400K steps.}
\resizebox{\linewidth}{!}{
\begin{tabular}{@{}lrrrrr@{}}
\toprule
\textbf{Window Size $W_s$} & \textbf{FID} $\downarrow$ & \textbf{sFID} $\downarrow$ & \textbf{IS} $\uparrow$ & \textbf{Prec.} $\uparrow$ & \textbf{Rec.} $\uparrow$ \\
\midrule
$W_s = 4$   & 30.16    & \textbf{8.77}    & 53.10    & 0.551    & \textbf{0.626}    \\
$W_s = 8$   & \textbf{29.89} & 8.78 & 53.28 & 0.550 & 0.623 \\
$W_s = 16$  & 29.97    & 8.93    & \textbf{54.02}    & \textbf{0.556}    & 0.618    \\
\bottomrule
\end{tabular}
}
\label{tab_winsize}
\end{table}

As shown in Table~\ref{tab_winsize}, when the window size is too small ($W_s = 4$), each window contains only $4\times4 = 16$ tokens, which limits the receptive field of the attention branch and reduces its ability to capture sufficient local context. Conversely, when the window size is too large ($W_s = 16$), the attention computation approaches global attention, reintroducing the computational redundancy that PSWA is designed to avoid. The setting $W_s = 8$ strikes a practical balance: it covers a sufficiently large local region to capture the dominant attention distances observed in Figure~\ref{fig2} (mostly within 2--6 tokens), while keeping the computational cost well below that of full attention.

In addition to the window-size analysis, we study the local neighborhood range of the high-frequency bridging branch through the Kth-order similarity ablation in Table~\ref{tab8}, where the DWConv kernel size is $2K-1$. This ablation evaluates how the Kth-order neighborhood range affects generation quality.

To explore whether the Kth-order attention is truly effective, we conduct ablation experiments on PiT-B at the channel level. All the experiments are trained for 200K iterations on ImageNet-1K. In Table~\ref{tab7}, \(C_{win}\) denotes the channels in the static window attention branch, and \(C_{bridge}\) represents the high-frequency bridging branch channels. The first row of Table~\ref{tab7} indicates that the model without the core static window attention and only with the bridging branch is the worst, with a FID of merely 29.15. The second row shows that the FID with only static window attention and no bridging branch is still only 23.29.
\begin{table}[!htbp]
\centering
\caption{Ablation study on ImageNet evaluating channel allocation strategies for the static window attention and high-frequency bridging branches, where all models were trained for 200K iterations.}
\resizebox{\linewidth}{!}{
\begin{tabular}{@{}lrrrrr@{}}
\toprule
\textbf{PiT-B}               & \textbf{FID} \(\downarrow\) & \textbf{sFID} \(\downarrow\) & \textbf{IS} \(\uparrow\) & \textbf{Prec.} \(\uparrow\) & \textbf{Rec.} \(\uparrow\) \\
\midrule
\(C_{win} = 0, C_{bridge} = C\)            & 29.15                   & 7.51                      & 58.46                  & 0.610                    & 0.607        \\
\(C_{win} = C, C_{bridge} = 0\)             & 23.29                   & 6.88                      & 65.21                  & 0.608                    & 0.610        \\
\(C_{win} / C \downarrow, C_{bridge} / C \uparrow\)                  & 22.92                     & 6.58                      & 66.37                  & 0.613                    & 0.619       \\
\(C_{win} / C=C_{bridge} / C\)            & 22.83                     & 6.51                      & 66.38                  & 0.609                    & 0.616       \\
\(C_{win} / C \uparrow, C_{bridge} / C \downarrow\)           & \textbf{21.80}                     & \textbf{6.34}                      & \textbf{67.8}                  & \textbf{0.625}                    & \textbf{0.626}            \\
\bottomrule
\end{tabular}
}
\label{tab7}
\end{table}
The third row reveals that when the static window attention branch channels gradually decrease and the high-frequency bridging branch channels gradually increase (without absorbing the Kth-order attention), the FID is 22.92, which is still not very high. The fourth row demonstrates that when the channels of the two branches are divided equally, the result is only slightly better, with an FID of 22.83. The fifth row indicates that when the static window attention branch channels gradually increase and the high-frequency bridging branch channels gradually decrease (fully absorb the information of the Kth-order neighborhood), achieving the best performance with an FID reaching 21.80.

We explore the influence of the range of K for the Kth-order attention, that is, the influence of the DWConv kernel-size (where kernel-size = 2K - 1). As shown in Table~\ref{tab8}, when K=1, the model uses only first-order attention, resulting in the worst performance with an FID score of 17.53. Both K = 2 and K = 5 achieve the optimal FID value of 15.56. The Precision (Prec.) metric reaches its highest value of 0.656 when K = 3, while the values for the other three cases are also comparable. Furthermore, the optimal values for the three metrics, namely sFID, Inception Score (IS), and Recall (Rec.), are all obtained when K = 5.

\begin{table}[!htbp]
\centering
\caption{Ablation study on the range K of Kth-order similarity. All results were trained on ImageNet-1K for 400K steps.}
\resizebox{\linewidth}{!}{
\begin{tabular}{@{}lrrrrr@{}}
\toprule
\textbf{PiT-B}              & \textbf{FID} $\downarrow$ & \textbf{sFID} $\downarrow$ & \textbf{IS} $\uparrow$ & \textbf{Prec.} $\uparrow$ & \textbf{Rec.} $\uparrow$ \\
\midrule
1st-order Attn            & 17.53	&6.56	&86.34	&0.637	&0.632                \\
2nd-order Attn              &\textbf{15.56}	&6.02	&88.48	&0.648	&0.631                \\
3rd-order Attn                  &15.69	&6.03	&88.03	&\textbf{0.656}	&0.636                 \\
4th-order Attn             &15.80	&6.05	&87.7	&0.648	&0.636                  \\
5th-order Attn            &\textbf{15.56}	 &\textbf{6.00}	&\textbf{88.86}	&0.648	&\textbf{0.647}                   \\
6th-order Attn             &15.92	&6.08	&87.94	&0.645	&0.632   \\
\bottomrule
\end{tabular}
}
\label{tab8}
\end{table}

\subsection{Visualization}
\begin{figure}
 \centering
 \includegraphics[width=\linewidth]{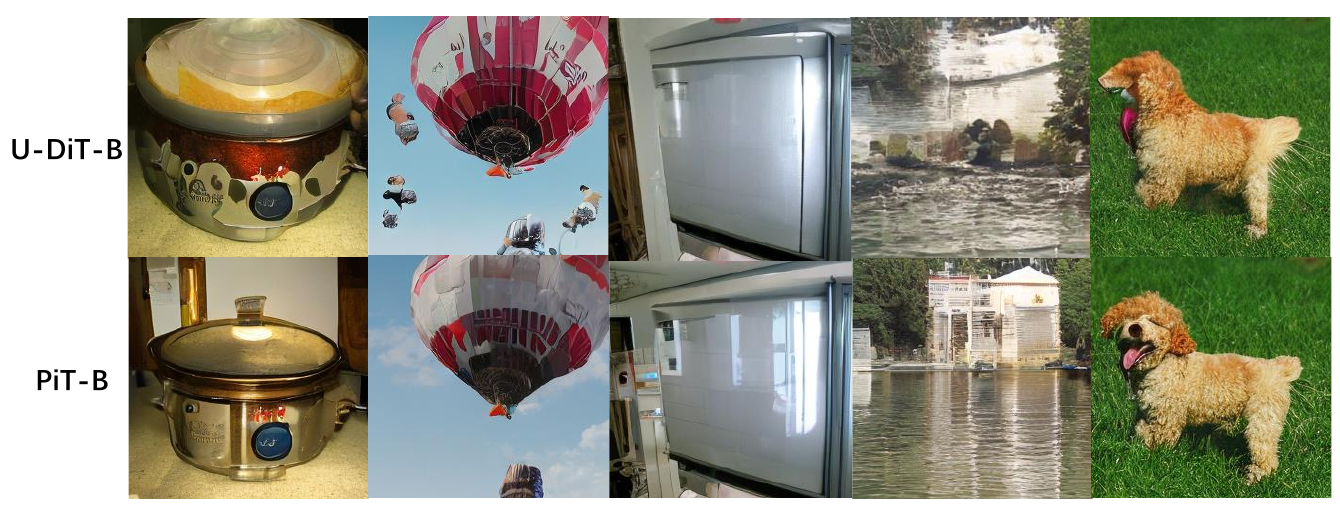}
 \caption{The comparison of image generation results among U-DiT-B and PiT-B models. Owing to the high-frequency information provided by the bridging branch, PiT-B exhibits superior detail and texture. All images are from the ImageNet-1K dataset~\cite{deng2009imagenet}}
 \label{fig7}
\end{figure}

We visualize the results of U-DiT-B and PiT-B, with all models trained on ImageNet-1K for 400K iterations under the same training settings. As illustrated in Figure~\ref{fig7}, PiT-B evidently exhibits superior detail and texture. For instance, in the first column, the generated image of the electric cooker by U-DiT-B appears blurred and indistinct. In contrast, PiT-B produces images with sharp and exquisite details. 

\begin{figure}
\centering
 {
 \includegraphics[width=\linewidth]{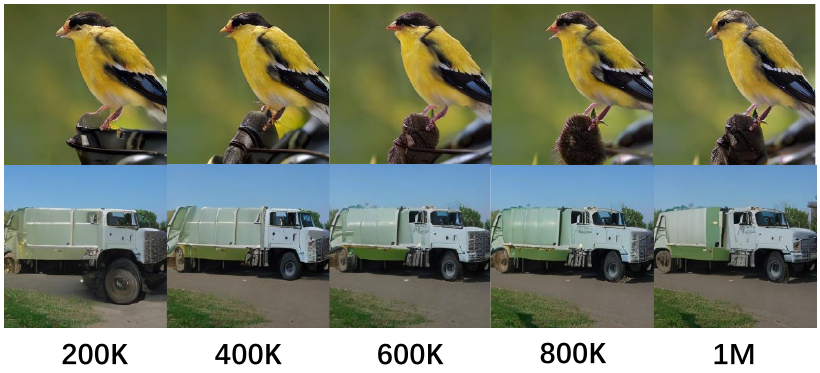}}
	\caption{As training iterations increase, the generated images of PiT-L are richer in textural details. All images are from the ImageNet-1K~\cite{deng2009imagenet}}
	\label{fig8}
\end{figure}

Figure~\ref{fig8} shows the process where the results generated by PiT-L get better and better as the training steps increase. We also supply visualizations Figure~\ref{sfig1} and Figure~\ref{sfig2}. All of them are the inference results of PiT-B after being trained for 400K steps at a $512\times512$ resolution.

\subsection{Generalization to Video Generation Tasks}
\label{sec:video_generalization}

We further evaluate PiT on the image-to-dance video generation task using the public Fashion\_video dataset. The dataset contains human turning and dancing sequences with changes in body pose and motion. In this experiment, the proposed PSWA/PCCA design is used as the Transformer denoising block in the PiT video variant. Following the evaluation setting of Wan-Animate~\cite{cheng2025wananimate}, we compare the PiT video variant with Wan-Animate using Fréchet Video Distance (FVD), where a lower value indicates better video generation quality.

\begin{table}[h]
\centering
\caption{Comparison of image-to-dance video generation on the Fashion\_video dataset.}
\label{tab:video_fvd}
\begin{tabular}{@{}lr@{}}
\toprule
\textbf{Method} & \textbf{FVD} $\downarrow$ \\
\midrule
Wan-Animate & 128.47 \\
PiT video variant (Ours) & \textbf{102.14} \\
\bottomrule
\end{tabular}
\end{table}
As shown in Table~\ref{tab:video_fvd}, the PiT video variant achieves an FVD of 102.14, while Wan-Animate obtains an FVD of 128.47. PiT reduces the FVD by 26.33 points. The quantitative results show that the proposed PSWA/PCCA design remains effective when applied to the video generation task.

\begin{figure}
\centering
\includegraphics[width=\linewidth]{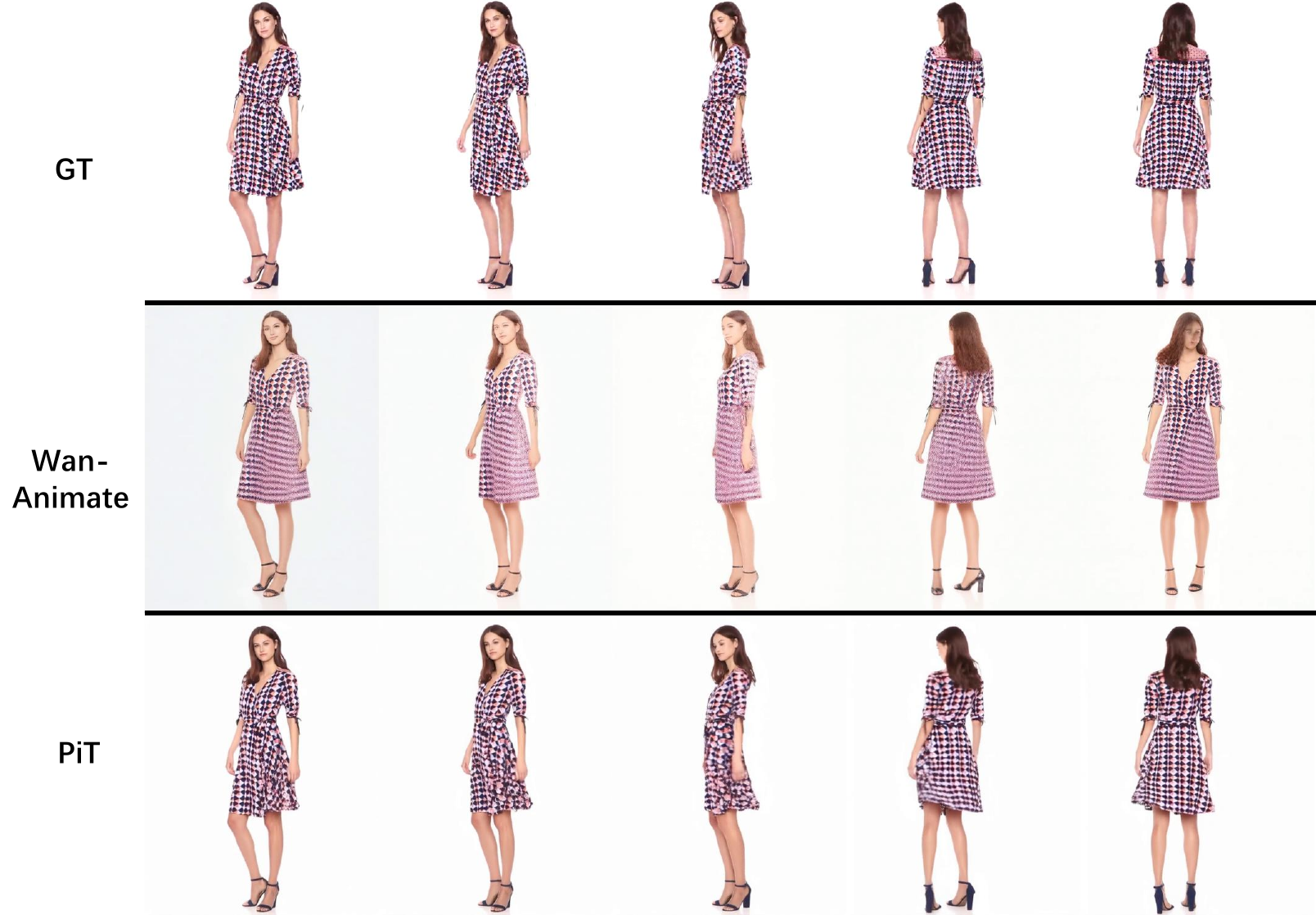}
\caption{Visual comparison of Wan-Animate and the PiT video variant on the image-to-dance task.}
\label{dance_compare}
\end{figure}

Figure~\ref{dance_compare} presents the visual comparison between the two methods. Compared with Wan-Animate, the PiT video variant preserves more clothing texture details and generates clearer head regions in the tested image-to-dance sequences. The quantitative and visual results provide an initial evaluation of PiT beyond image generation. Since the current experiment focuses on one image-to-dance dataset, evaluation on more video generation tasks will be considered in future work.

\section{Limitations and Failure Cases}
\label{sec:limitations}
\begin{figure}
\centering
\includegraphics[width=\linewidth]{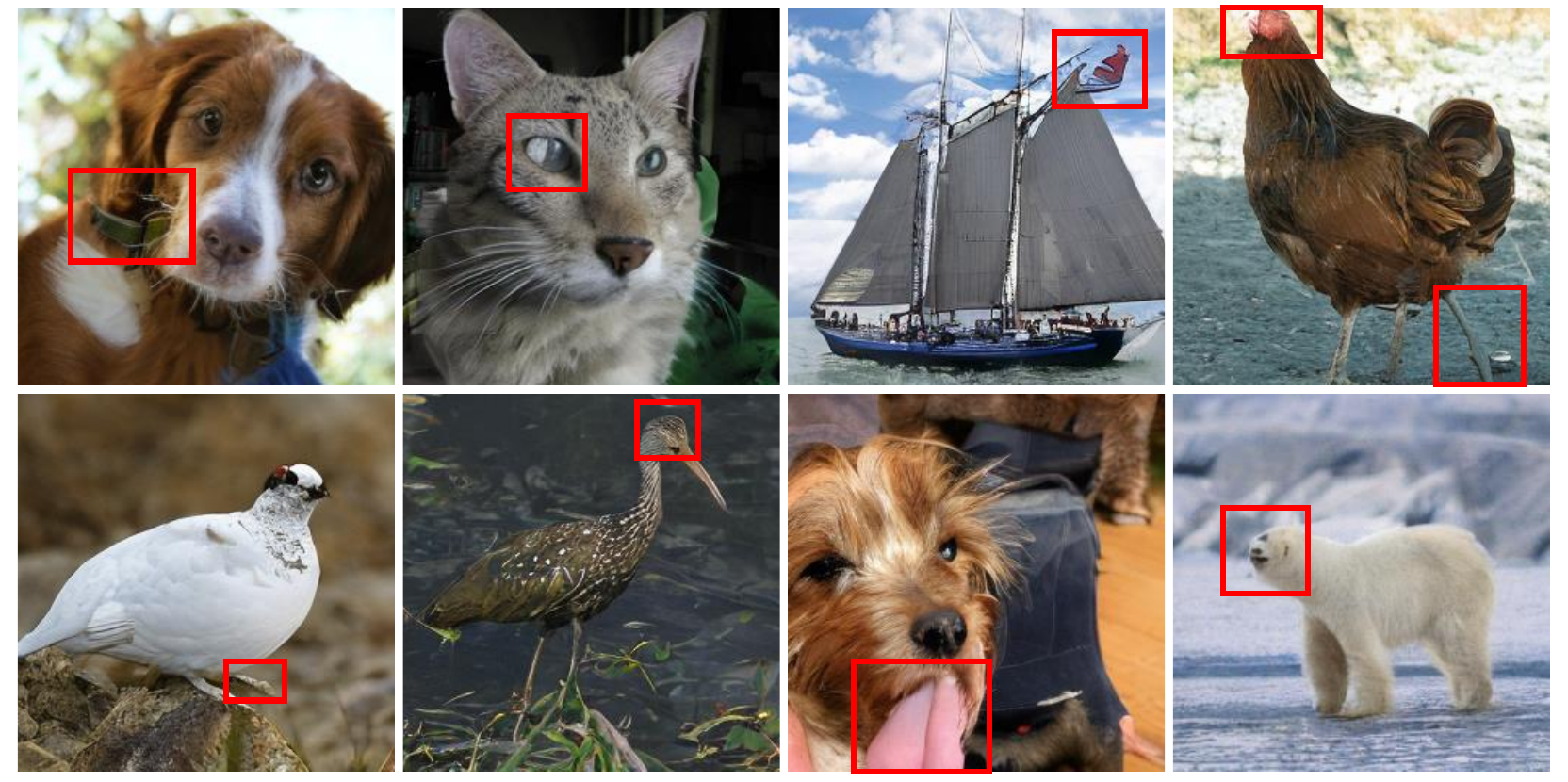}
\caption{Qualitative examples of failed PiT generations.}
\label{failure}
\end{figure}

\begin{figure}
\centering
\includegraphics[width=0.92\linewidth]{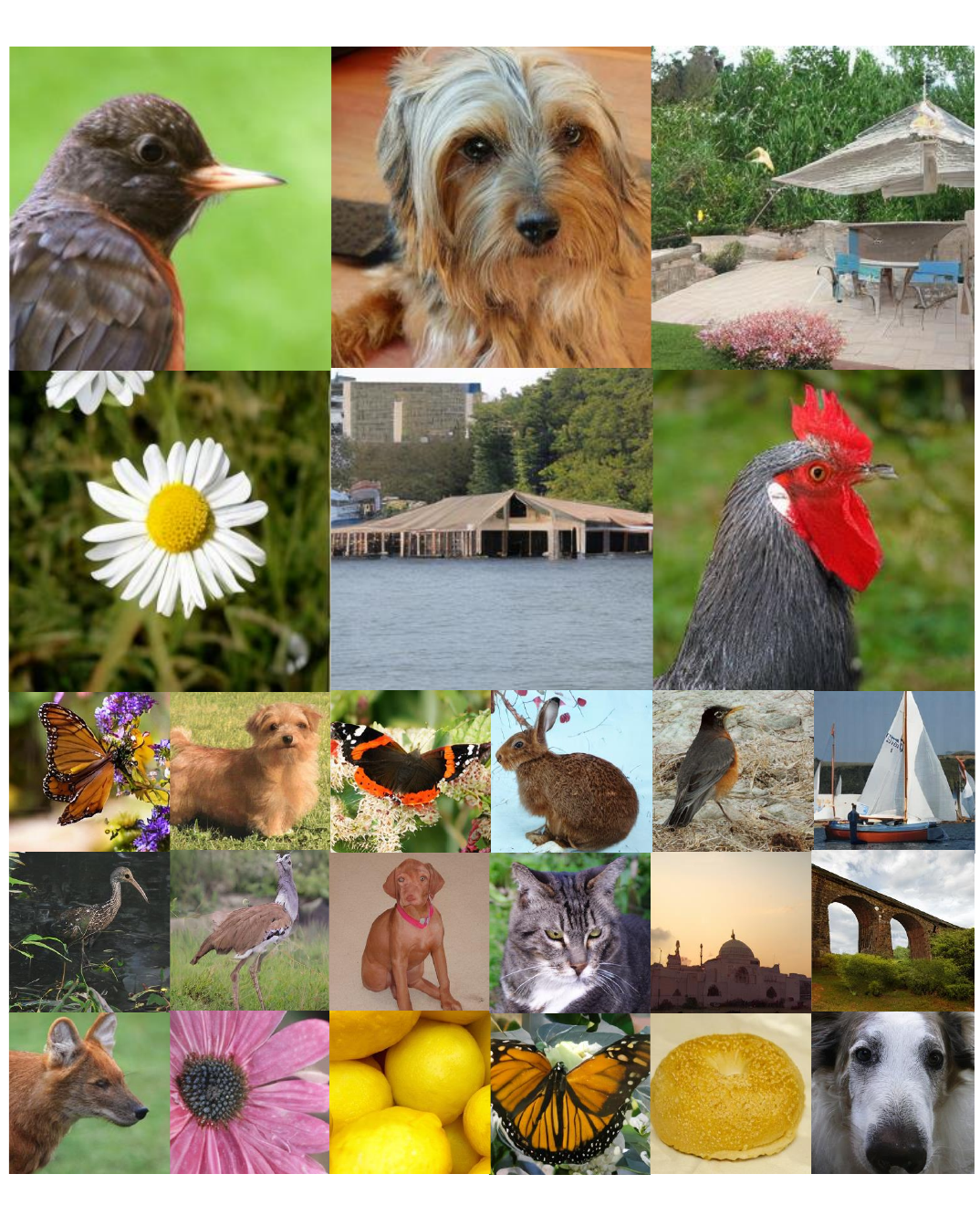}
\caption{Supplementary visualization results 1.}
\label{sfig1}
\end{figure}

\begin{figure}
\centering
\includegraphics[width=0.92\linewidth]{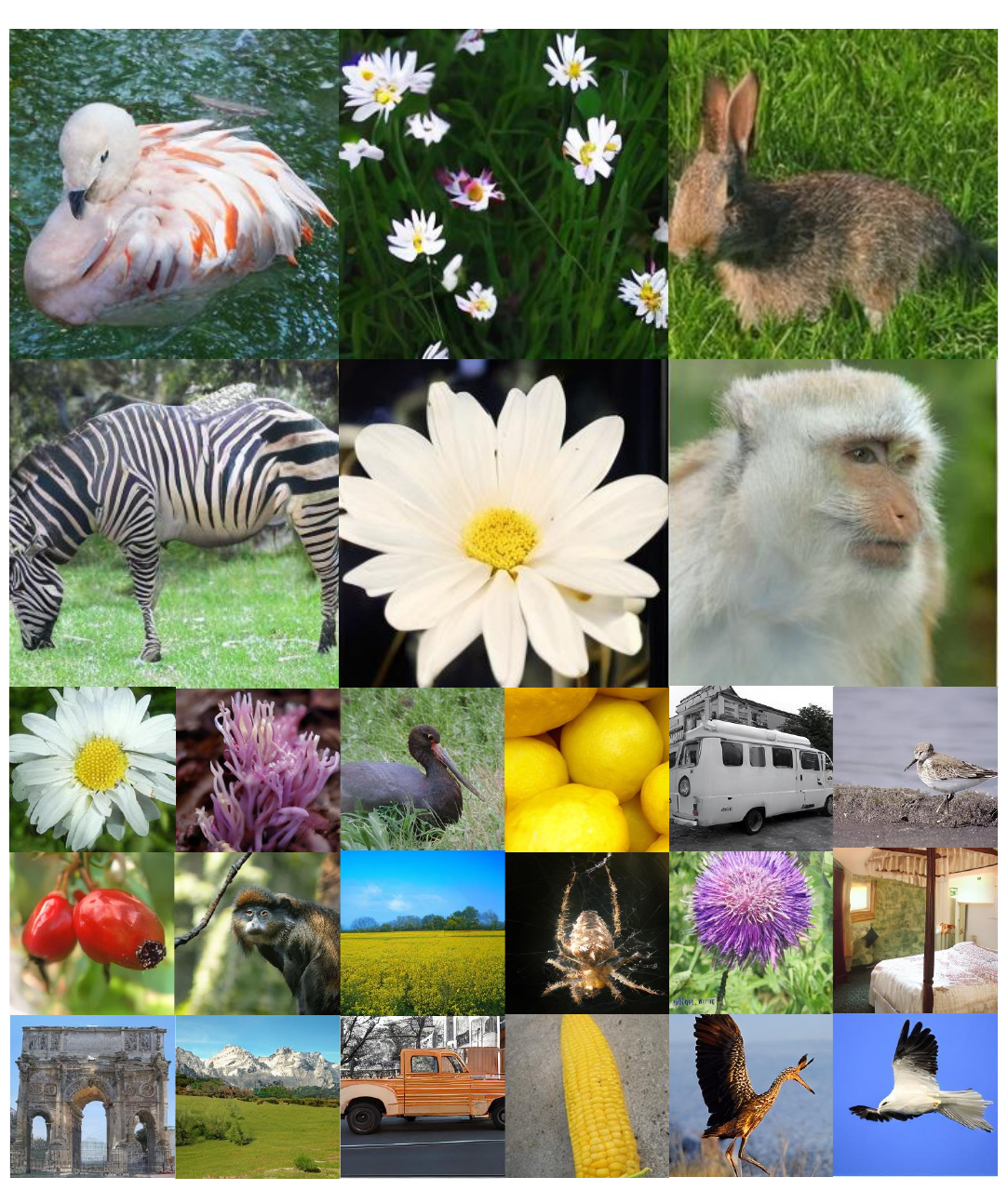}
\caption{Supplementary visualization results 2.}
\label{sfig2}
\end{figure}

Although PiT improves efficiency and representation quality, it still has several limitations. First, the window size is fixed across layers and timesteps, which may be suboptimal because different denoising stages can require different dependency ranges. Second, the PCCA allocation is manually designed rather than learned automatically, leaving room for adaptive channel-allocation strategies. Third, restricting attention to local windows may theoretically weaken long-range scene coherence in images that require strong global structure. This risk is mitigated by the U-shaped bottleneck, where the spatial resolution is reduced to $8\times8$ and the default window size covers the entire feature map, and by the DWConv branch, which continuously bridges neighboring windows. We show several failure cases in Fig.~\ref{failure}. These failures mainly involve insufficient generation quality in fine-grained animal regions, such as the eyes, mouths, and feet. This motivates future work on adaptive windows, learnable channel allocation, and stronger global context modeling. Broader validation on text-to-image generation and more diverse video benchmarks also remains an important direction for future work.

\section{Conclusion}
In this paper, we propose PiT, a progressive diffusion transformer that improves both computational efficiency and representation quality. The proposed PSWA alleviates redundant long-distance attention and compensates for the low-frequency inertia of standard attention through a parallel high-frequency bridging branch. We further introduce Kth-order attention and PCCA, which captures higher-order similarity by progressively reallocating the existing channel budget between attention and DWConv branches, without introducing additional parameters or FLOPs beyond the base architecture. Extensive quantitative, ablation, efficiency, and visualization results demonstrate the effectiveness and scalability of PiT under the evaluated settings.

\bibliographystyle{elsarticle-num}
\bibliography{refs}





\end{document}